\documentclass[11pt,a4paper]{article}
\usepackage[hyperref]{emnlp2020}
\usepackage{times}
\usepackage{latexsym}
\usepackage{graphicx}

\usepackage{xcolor}

\usepackage{microtype}

\aclfinalcopy % Uncomment this line for the final submission
\def\aclpaperid{1997} %  Enter the acl Paper ID here

\title{Multilingual Offensive Language Identification \\ with Cross-lingual Embeddings}

\author{Tharindu Ranasinghe \\
  University of Wolverhampton \\
  Wolverhampton, UK\\
  \texttt{tharindu.ranasinghe@wlv.ac.uk} \\\And
  Marcos Zampieri \\
  Rochester Institute of Technology \\
  Rochester, NY, USA \\
  \texttt{mazgla@rit.edu} \\}

\date{}

\begin{document}
\maketitle
\begin{abstract}
Offensive content is pervasive in social media and a reason for concern to companies and government organizations. Several studies have been recently published investigating methods to detect the various forms of such content (e.g. hate speech, cyberbulling, and cyberaggression). The clear majority of these studies deal with English partially because most annotated datasets available contain English data. In this paper, we take advantage of English data available by applying cross-lingual contextual word embeddings and transfer learning to make predictions in languages with less resources. We project predictions on comparable data in Bengali, Hindi, and Spanish and we report results of 0.8415 F1 macro for Bengali, 0.8568 F1 macro for Hindi, and 0.7513 F1 macro for Spanish. Finally, we show that our approach compares favorably to the best systems submitted to recent shared tasks on these three languages, confirming the robustness of cross-lingual contextual embeddings and transfer learning for this task. 
\end{abstract}

\section{Introduction}

Offensive posts on social media result in a number of undesired consequences to users. They have been investigated as triggers of suicide attempts and ideation, and mental health problems \cite{bonanno2013cyber,bannink2014cyber}. One of the most common ways to cope with offensive content online is training systems to be capable of recognizing offensive messages or posts. Once recognized, such offensive content can be set aside for human moderation or deleted from the respective platform (e.g. Facebook, Twitter), preventing harm to users and controlling the spread of abusive behavior in social media. 

There have been several recent studies published on automatically identifying various kinds of offensive content such as abuse \cite{mubarak2017}, aggression \cite{kumar2018benchmarking,trac2020}, cyber-bullying \cite{rosa2019automatic}, and hate speech \cite{malmasi2018}. While there are a few studies published on languages such as Arabic \cite{mubarak2020arabic} and Greek \cite{pitenis2020}, most studies and datasets created so far include English data. Data augmentation \cite{ghadery2020liir} and multilingual word embeddings \cite{pamungkas2019cross} have been applied to take advantage of existing English resources to improve the performance in systems dealing with languages other than English. To the best of our knowledge, however, state-of-the-art cross-lingual contextual embeddings such as XLM-R \cite{conneau2019unsupervised} have not yet been applied to offensive language identification. To address this gap, we evaluate the performance of cross-lingual contextual embeddings and transfer learning (TL) methods in projecting predictions from English to other languages. We show that our methods compare favorably to state-of-the-art approaches submitted to recent shared tasks on all datasets. The main contributions of this paper are the following:

\begin{enumerate}
    \item We apply cross-lingual contextual word embeddings to offensive language identification. We take advantage of existing English data to project predictions in three other languages: Bengali, Hindi, and Spanish.
    \vspace{-2mm}
    \item We tackle both off-domain and off-task data for Bengali. We show that not only can these methods project predictions for different languages but also for different domains (e.g. Twitter vs. Facebook) and tasks (e.g. binary vs. three-way classification).
    \vspace{-2mm}
    \item We provide important resources to the community: the code, and the English model will be freely available to everyone interested in working on low-resource languages using the same methodology.
\end{enumerate}

\section{Related Work}
\label{sec:RW}

There is a growing interest in the development of computational models to identify offensive content online. Early approaches relied heavily on feature engineering combined with traditional machine learning classifiers such as naive bayes and support vector machines \cite{xu2012learning, dadvar2013improving}. More recently, neural networks such as LSTMs, bidirectional LSTMs, and GRUs combined with word embeddings have proved to outperform traditional machine learning methods in this task \cite{aroyehun2018aggression, majumder2018filtering}. In the last couple of years, transformer models like ELMO \cite{peters-etal-2018-deep} and BERT \cite{devlin2019bert} have been applied to offensive language identification achieving competitive scores and topping the leaderboards in recent shared tasks \cite{liu-etal-2019-nuli,ranasinghe2019brums}. Most of these approaches use existing pre-trained transformer models which can also be used as text classification models.

The clear majority of studies on this topic deal with English \cite{malmasi2017detecting,yao2019cyberbullying,ridenhour2020detecting} partially motivated by the availability English resources (e.g. corpora, lexicon, and pre-trained models). In recent years, a number of studies have been published on other languages such as Arabic \cite{mubarak2020arabic}, Danish \cite{sigurbergsson2020offensive}, Dutch \cite{tulkens2016dictionary}, French \cite{chiril-etal-2019-multilingual}, Greek \cite{pitenis2020}, Italian \cite{poletto2017hate}, Portuguese \cite{fortuna2019hierarchically}, Slovene \cite{fiser2017}, and Turkish \cite{coltekin2020} creating new datasets and resources for these languages.

Recent competitions organized in 2020 such as TRAC \cite{trac2020} and OffensEval \cite{zampieri2020semeval} have included datasets in multiple languages providing participants with the opportunity to explore cross-lingual learning models opening exciting new avenues for research on languages other than English and, in particular, on low-resource languages. The aforementioned deep learning methods require large annotated datasets to perform well which is not always available for low-resource languages. In this paper, we address the problem of data scarcity in offensive language identification by using transfer learning and cross-lingual transformers from a resource rich language like English to three other languages: Bengali, Hindi, and Spanish.

\section{Data}
\label{sec:data}

We acquired datasets in English and three other languages: Bengali, Hindi, and Spanish (listed in Table \ref{tab:data}). The four datasets have been used in shared tasks in 2019 and 2020 allowing us to compare the performance of our methods to other approaches. As our English dataset, we chose the Offensive Language Identification Dataset (OLID) \cite{OLID}, used in the SemEval-2019 Task 6 (OffensEval) \cite{offenseval}. OLID is arguably one of the most popular offensive language datasets. It contains manually annotated tweets with the following three-level taxonomy and labels:

\vspace{-2mm}

\begin{itemize}
%[noitemsep,topsep=3pt]%,leftmargin=*]
    \item[\bf A:] Offensive language identification - offensive vs. non-offensive;
    \vspace{-2mm}
    \item[\bf B:] Categorization of offensive language - targeted insult or thread vs. untargeted profanity;
    \vspace{-6mm}
    \item[\bf C:] Offensive language target identification - individual vs. group vs. other.
\end{itemize}

\vspace{-2mm}

\noindent We chose OLID due to the flexibility provided by its hierarchical annotation model that considers multiple types of offensive content in a single taxonomy (e.g. targeted insults to a group are often {\em hate speech} whereas targeted insults to an individual are often {\em cyberbulling}). This allows us to map OLID level A (offensive vs. non-offensive) to labels in the other three datasets. OLID's annotation model is intended to serve as a general-purpose model for multiple {\em abusive language detection sub-tasks}  \cite{waseem2017understanding}. The transfer learning strategy used in this paper provides us with an interesting opportunity to evaluate how closely the OLID labels relate to the classes in datasets annotated using different guidelines and sub-task definitions (e.g. {\em aggression} and {\em hate speech}).
%Other tasks such as HatEval and HASOC 2019. In this paper we take a advantage ot OLID's first layer of annotation and mapping it to comparable datasets.

%\vspace{-2mm}

% The Bengali dataset \cite{trac2-dataset} was used in the TRAC-2 shared task \cite{trac2020} on aggression identification. It contains 5,000 Facebook posts annotated with three labels: overtly aggressive, covertly aggressive, and non aggressive. It is different than the other three datasets in terms of domain (Facebook instead of Twitter) and set of labels (three classes instead of binary) allowing us to compare the performance of cross-lingual embeddings on off-domain data and off-task data. The Hindi dataset \cite{hasoc2019} contains 8,000 tweets and it was used in the HASOC 2019 shared task. The annotation features two categories: Non Hate-Offensive and Hate-Offensive which is roughly equivalent to OLID's level A annotation. The Spanish dataset \cite{hateval2019} was used in SemEval-2019 Task 5 (HatEval). It contains a total of 6,600 tweets targetted at women and migrants annotated as hateful or non-hateful making it similar to OLID's level A annotation. 

The Hindi dataset \cite{hasoc2019} was used in the HASOC 2019 shared task, while the Spanish dataset \cite{hateval2019} was used in SemEval-2019 Task 5 (HatEval). They both contain Twitter data and two labels. The Bengali dataset \cite{trac2-dataset} was used in the TRAC-2 shared task \cite{trac2020} on aggression identification. It is different than the other three datasets in terms of domain (Facebook instead of Twitter) and set of labels (three classes instead of binary), allowing us to compare the performance of cross-lingual embeddings on off-domain data and off-task data. 
%It contains a total of 6,600 tweets targetted at women and migrants annotated as hateful or non-hateful making it similar to OLID's level A annotation. 

\begin{table}[!ht]
\centering
\setlength{\tabcolsep}{4.5pt}
\scalebox{.85}{
\begin{tabular}{lccp{5.2cm}}
\hline
\bf Lang. & \bf Inst. & \bf S & \bf Labels  \\ \hline

Bengali & 4,000 & F & overtly aggressive, covertly aggressive, non aggressive     \\
English & 14,100 & T & offensive, non-offensive \\
Hindi & 8,000 & T & hate offensive, non hate-offensive \\
Spanish & 6,600 & T & hateful, non-hateful      \\
\hline
\end{tabular}
}
\caption{Instances (Inst.), source (S) and labels in all datasets. F stands for Facebook and T for Twitter.}
\label{tab:data}
\end{table}

\section{Methodology}
%In this section we will describe the methodology of our experiment. 

\begin{figure}[!ht]
\centering
\includegraphics[scale=0.72]{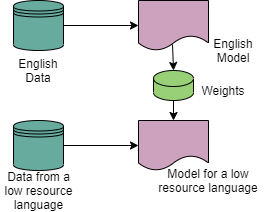}
\caption{Transfer learning strategy.}
\label{fig:transfer_learning}
\end{figure}

Transformer models have been used successfully for various NLP tasks \cite{devlin2019bert}. Most of the tasks were focused on English language due to the fact the most of the pre-trained transformer models were trained on English data. Even though, there were several multilingual models like BERT-m \cite{devlin2019bert} there was much speculations about its ability to represent all the languages \cite{pires-etal-2019-multilingual} and although BERT-m model showed some cross-lingual characteristics it has not been trained on crosslingual data \cite{karthikeyan2020cross}. The motivation behind this methodology was the recently released cross-lingual transformer models - XLM-R \cite{conneau2019unsupervised} which has been trained on 104 languages. The interesting fact about XLM-R is that it is very compatible in monolingual benchmarks while achieving the best results in cross-lingual benchmarks at the same time \cite{conneau2019unsupervised}. The main idea of the methodology is that we train a classification model on a resource rich, typically English, using a cross-lingual transformer model, save the weights of the model and when we initialise the training process for a lower resource language, start with the saved weights from English. This process is also known as transfer learning and is illustrated in Figure \ref{fig:transfer_learning}.

\noindent There are two main parts of the methodology. Subsection \ref{subsec:classification} describes the classification architecture we used for all the languages. In Subsection \ref{subsec:transfer} we describe the transfer learning strategies used to take advantage of English offensive language data in predicting offense in less-resourced languages.

\subsection{XLM-R for Text Classification}
\label{subsec:classification}
Similar to other transformer architectures XLM-R transformer architecture can also be used for text classification tasks \cite{conneau2019unsupervised}. XLM-R-large model contains approximately 125M parameters with 12-layers, 768 hidden-states, 3072 feed-forward hidden-states and 8-heads \cite{conneau2019unsupervised}. It takes an input of a sequence of no more than 512 tokens and outputs the representation of the sequence. The first token of the sequence is always [CLS] which contains
the special classification embedding \cite{10.1007/978-3-030-32381-3_16}.

For text classification tasks, XLM-R takes the final hidden state \textbf{h} of the first token [CLS] as the representation of the whole sequence. A simple softmax classifier is added to the top of XLM-R to predict the probability of label c: as shown in Equation \ref{equ:softmax} where W is the task-specific parameter matrix.

\begin{equation}
\label{equ:softmax}
p(c|\textbf{h}) = softmax(W\textbf{h}) 
\end{equation}

\begin{figure}[ht]
\centering
\includegraphics[scale=0.4]{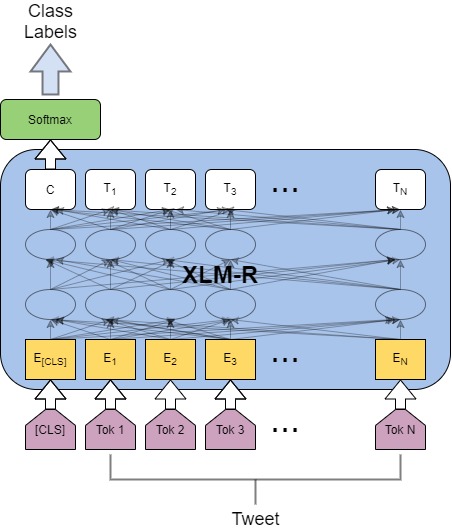}
\caption{Text classification architecture with XLM-R.}
\label{fig:architecture}
\end{figure}

\noindent We fine-tune all the parameters from XLM-R as well as W jointly by maximising the log-probability of the correct label. The architecture diagram of the classification is shown in Figure \ref{fig:architecture}. We specifically used the XLM-R large model.

\subsection{Transfer-learning strategies}
\label{subsec:transfer}
When we adopt XLM-R for multilingual offensive language identification, we perform transfer learning in two different ways. 

\paragraph{Inter-language transfer learning} We first trained the XLM-R classification model on first level of English offensive language identification dataset (OLID) \cite{OLID}. Then we save the weights of the XLM-R model as well as the softmax layer. We use these saved weights from English to initialise the weights for a new language. To explore this transfer learning aspect we experimented on Hindi language which was released for HASOC 2019 shared task \cite{hasoc2019} and on Spanish data released for Hateval 2019 \cite{hateval2019}.
  
\paragraph{Inter-task and inter-language transfer learning} Similar to the inter-language transfer learning strategy, we first trained the XLM-R classification model on the first level of English offensive language identification dataset (OLID) \cite{OLID}. Then we only save the weights of the XLM-R model and use the saved weights to initialise the weights for a new language. We did not use the weights of the last softmax layer since we wanted to test this strategy on data that has a different number of offensive classes to predict. We explored this transfer learning aspect with  Bengali dataset released with TRAC - 2 shared task \cite{trac2020}. As described in the Section \ref{sec:data} the classifier should make a 3-way classification in between ‘Overtly Aggressive’, ‘Covertly Aggressive’ and ‘Non Aggressive’ text data.

We used a Nvidia Tesla K80 GPU to train the models. We divided the dataset into a training set and a validation set using 0.8:0.2 split on the dataset. We predominantly fine tuned the learning rate and number of epochs of the classification model manually to obtain the best results for the validation set. We obtained $1e^-5$ as the best value for learning rate and 3 as the best value for number of epochs for all the languages. Training for English language took around 1 hour while training for other languages took around 30 minutes. The code and the pretrained English model is available on GitHub.\footnote{The public GitHub repository are available on \url{https://github.com/tharindudr/DeepOffense}}

\section{Results and Evaluation}

We evaluate the results obtained by all models using the test sets provided by the organizers of each competition. We compared our results to the best systems in TRAC-2 for Bengali, HASOC for Hindi, HatEval for Spanish in terms of weighted and macro F1 score according to the metrics reported by the task organizers - TRAC-2 reported only macro F1, HatEval reported only weighted F1, and HASOC reported both. 

Finally, we evaluate the improvement of the transfer learning strategy in the performance of both BERT and XLM-R. We present the results along with the majority class baseline for each language in Table \ref{tab:all}. \textit{TL} indicates that the model used the inter language transfer learning strategy described in Subsection \ref{subsec:transfer}. 

\begin{table}[!ht]
\centering
\setlength{\tabcolsep}{4.5pt}
\scalebox{0.85}{
\begin{tabular}{llcc}
\hline
\bf Language & \bf Model   & \bf M F1 & \bf W F1 \\ \hline

& XLM-R (TL)  & 0.8415   & 0.8423     \\
& \newcite{risch2020bagging} & 0.8219 & \\
Bengali & BERT-m (TL) & 0.8197   & 0.8231      \\
& XLM-R       & 0.8142   & 0.8188      \\
& BERT-m      & 0.8132   & 0.8157      \\
& Baseline & 0.2498 &	0.4491 \\
 \hline

& XLM-R (TL)  & 0.8568   & 0.8580     \\
& BERT-m (TL) & 0.8211   & 0.8220      \\
Hindi & \newcite{bashar2019qutnocturnal} & 0.8149 & 0.8202 \\
& XLM-R       & 0.8061   & 0.8072      \\
& BERT-m      & 0.8025   & 0.8030      \\
& Baseline & 0.3510 &	0.3798 \\
 \hline
 
& XLM-R (TL)  & 0.7513   & 0.7591     \\ 
& BERT-m (TL) & 0.7319   & 0.7385      \\
& \newcite{vega2019mineriaunam} & & 0.7300 \\
Spanish & \newcite{perez2019atalaya} & & 0.7300 \\
& XLM-R       & 0.7224   & 0.7265      \\
& BERT-m      & 0.7215   & 0.7234      \\
& Baseline & 0.3700 &	0.4348 \\
\hline
\end{tabular}
}
\caption{Results ordered by macro (M) F1 for Bengali and weighted (W) F1 for Hindi and Spanish.}
\label{tab:all}
\end{table}

%Table \ref{tab:hindi} presents the results of the several models on Hindi. 
%\textit{TL} indicates that the model used the inter language transfer learning strategy described in Subsection \ref{subsec:transfer}. 

\noindent For Hindi, transfer learning with XLM-R cross lingual embeddings provided the best results achieving 0.8568 and 0.8580 weighted and macro F1 score respectively. In HASOC 2019 \cite{hasoc2019}, the best model by \newcite{bashar2019qutnocturnal} scored 0.8149 Macro F1 and 0.8202 Weighted F1 using convolutional neural networks.  

%The XLM-R with inter language transfer learning strategy outperforms all other models in both metrics.

% \begin{table}[!ht]
% \scalebox{0.85}{
% \begin{tabular}{lcc}
% \hline
% \bf Model   & \bf Macro F1 & \bf Weighted F1 \\ \hline
% XLM-R (TL)  & 0.8568   & 0.8580     \\
% BERT-m (TL) & 0.8211   & 0.8220      \\
% \newcite{bashar2019qutnocturnal} & 0.8149 & 0.8202 \\
% XLM-R       & 0.8061   & 0.8072      \\
% BERT-m      & 0.8025   & 0.8030      \\
%  \hline
% \end{tabular}
% }
% \caption{Results on Hindi ordered by weighted F1 score.}
% \label{tab:hindi}
% \end{table}

% \noindent Table \ref{tab:spanish} presents the results obtained for Spanish. Once again the results presented in \textit{TL} indicate that the model used the inter-language transfer learning strategy described in Subsection \ref{subsec:transfer}. 
For Spanish transfer learning with XLM-R cross lingual embeddings also provided the best results achieving 0.7513 and 0.7591 macro and weighted F1 score respectively. The best two models in HatEval \cite{hateval2019} for Spanish scored 0.7300 macro F1 score.
%\footnote{The HatEval organisers scored models using weighted F1 scores only.} 
Both models applied SVM classifiers trained on a variety of features like character and word n-grams, POS tags, offensive word lexicons, and embeddings. 

%However, our XLM-R model with inter-language transfer learning strategy outperforms the best model by a significant margin in macro F1 score.

% \begin{table}[!ht]
% \scalebox{0.85}{
% \begin{tabular}{lcc}
% \hline
% \bf Model   & \bf Macro F1 & \bf Weighted F1 \\ \hline
% XLM-R (TL)  & 0.7513   & 0.7591     \\ 
% BERT-m (TL) & 0.7319   & 0.7385      \\
% \newcite{vega2019mineriaunam} & & 0.7300 \\
% \newcite{perez2019atalaya} & & 0.7300 \\
% XLM-R       & 0.7224   & 0.7265      \\
% BERT-m      & 0.7215   & 0.7234      \\
% \hline
% \end{tabular}
% }
% \caption{Results on Spanish ordered by weighted F1 score.}
% \label{tab:spanish}
% \end{table}

%\noindent Table \ref{tab:bengali} shows the results we obtained for Bengali. 
The results for Bengali deserve special attention because the Bengali data is off-domain with respect to the English data (Facebook instead of Twitter), and it contains three labels (covertly aggressive, overtly aggressive, and not aggressive) instead of the two labels present in the English dataset (offensive and non-offensive). \textit{TL} indicates that the model used the inter-task, inter-domain, and inter-language transfer learning strategy described in Subsection \ref{subsec:transfer}. Similar to the Hindi and Spanish, transfer learning with XLM-R cross lingual embeddings provided the best results for Bengali achieving 0.7513 and 0.7591 macro and weighted F1 respectively thus outperforming the other models by a significant margin. The best model in the TRAC-2 shared task \cite{trac2020} scored 0.821 weighted F1 score in Bengali using a BERT-based system.
%\footnote{The TRAC-2 organisers scored models using macro F1 scores only.} 
%As can be seen in the table \ref{tab:bengali} XLM-R model with inter-task and inter-language transfer learning strategy outperforms the best model by a significant margin in weighted F1 score. 

\begin{figure}[ht]
\centering
\includegraphics[scale=0.62]{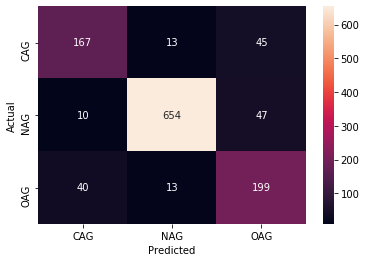}
\caption{Heat map of the Bengali test set predictions by XLM-R (TL).}
\label{fig:heatmap}
\end{figure}

We look closer at the test set predictions by XLM-R (TL) for Bengali in Figure \ref{fig:heatmap}. We observe that the performance for the non-aggressive class is substantially better than the performance for the overtly aggressive and covertly aggressive classes following a trend observed by the TRAC-2 participants including \newcite{risch2020bagging}. 
%The performance of XLM-R (TL) is higher for the covertly aggress \cite{risch2020bagging} 0.62 precision where 0.71. 

% \begin{table}[!ht]
% \scalebox{0.85}{
% \begin{tabular}{lcc}
% \hline
% \bf Model   & \bf Macro F1 & \bf Weighted F1 \\ \hline
% XLM-R (TL)  & 0.8415   & 0.8423     \\
% \newcite{risch2020bagging} & 0.8219 & \\
% BERT-m (TL) & 0.8197   & 0.8231      \\
% XLM-R       & 0.8142   & 0.8188      \\
% BERT-m      & 0.8132   & 0.8157      \\
%  \hline
% \end{tabular}
% }
% \caption{Results on Bengali ordered by macro F1 score.}
% \label{tab:bengali}
% \end{table}

\noindent Finally, it is clear that in all the experimental settings, the cross-lingual embedding models fine-tuned with transfer learning, outperforms the best system available for the three languages. Furthermore, the results show that the cross-lingual nature of the XLM-R model provided a boost over the multilingual BERT model in all languages tested. 

\section{Conclusion}

This paper is the first study to apply cross-lingual contextual word embeddings in offensive language identification projecting predictions from English to other languages using benchmarked datasets from shared tasks on Bengali \cite{trac2020}, Hindi \cite{hasoc2019}, and Spanish \cite{hateval2019}. We have showed that XLM-R with transfer learning outperforms all of the other methods we tested as well as the best results obtained by participants of the three competitions. 

The results obtained by our models confirm that OLID's general hierarchical annotation model encompasses multiple types of offensive content such as {\em aggression}, included in the Bengali dataset, and {\em hate speech} included in the Hindi and Spanish datasets, allowing us to model different sub-tasks jointly using the methods described in this paper. Furthermore, the results we obtained for Bengali show that it is possible to achieve high performance using transfer learning on off-domain (Twitter vs. Facebook) and off-task data when the labels do not have a direct correspondence in the projected dataset (two in English and three in Bengali). This opens exciting new avenues for future research considering the multitude of phenomena (e.g. {\em hate speech, aggression, cyberbulling}), annotation schemes and guidelines used in offensive language datasets.

In future work, we would like to further evaluate our models using SOLID, a novel large English dataset with over 9 million tweets \cite{rosenthal2020}, along with datasets in four other languages (Arabic, Danish, Greek, and Turkish) that were made available for the second edition of OffensEval \cite{zampieri2020semeval}. These datasets were collected using the same methodology and were annotated according to OLID's guidelines. Finally, we would also like to apply our models to languages with even less resources available to help coping with the problem of offensive language in social media. 

\section*{Acknowledgement}

We would like to thank the HASOC, HatEval, and OffensEval shared task organizers for making the datasets used in this paper available.

We further thank the anonymous EMNLP reviewers for providing us with valuable feedback to improve this paper.

\bibliographystyle{acl_natbib}
\bibliography{emnlp2020}

\begin{thebibliography}{45}
\expandafter\ifx\csname natexlab\endcsname\relax\def\natexlab#1{#1}\fi

\bibitem[{Aroyehun and Gelbukh(2018)}]{aroyehun2018aggression}
Segun~Taofeek Aroyehun and Alexander Gelbukh. 2018.
\newblock Aggression detection in social media: Using deep neural networks,
  data augmentation, and pseudo labeling.
\newblock In \emph{Proceedings of TRAC}.

\bibitem[{Bannink et~al.(2014)Bannink, Broeren, van~de Looij-Jansen, de~Waart,
  and Raat}]{bannink2014cyber}
Rienke Bannink, Suzanne Broeren, Petra~M van~de Looij-Jansen, Frouwkje~G
  de~Waart, and Hein Raat. 2014.
\newblock {Cyber and Traditional Bullying Victimization as a Risk Factor for
  Mental Health Problems and Suicidal Ideation in Adolescents}.
\newblock \emph{PloS one}, 9(4).

\bibitem[{Bashar and Nayak(2019)}]{bashar2019qutnocturnal}
Md~Abul Bashar and Richi Nayak. 2019.
\newblock {QutNocturnal@ HASOC'19: CNN for hate speech and offensive content
  identification in Hindi language}.
\newblock In \emph{Proceedings of FIRE}.

\bibitem[{Basile et~al.(2019)Basile, Bosco, Fersini, Nozza, Patti,
  Rangel~Pardo, Rosso, and Sanguinetti}]{hateval2019}
Valerio Basile, Cristina Bosco, Elisabetta Fersini, Debora Nozza, Viviana
  Patti, Francisco~Manuel Rangel~Pardo, Paolo Rosso, and Manuela Sanguinetti.
  2019.
\newblock {S}em{E}val-2019 task 5: Multilingual detection of hate speech
  against immigrants and women in twitter.
\newblock In \emph{Proceedings of SemEval}.

\bibitem[{Bhattacharya et~al.(2020)Bhattacharya, Singh, Kumar, Bansal, Bhagat,
  Dawer, Lahiri, and Ojha}]{trac2-dataset}
Shiladitya Bhattacharya, Siddharth Singh, Ritesh Kumar, Akanksha Bansal, Akash
  Bhagat, Yogesh Dawer, Bornini Lahiri, and Atul~Kr. Ojha. 2020.
\newblock Developing a multilingual annotated corpus of misogyny and
  aggression.
\newblock In \emph{Proceedings of TRAC}.

\bibitem[{Bonanno and Hymel(2013)}]{bonanno2013cyber}
Rina~A Bonanno and Shelley Hymel. 2013.
\newblock Cyber bullying and internalizing difficulties: Above and beyond the
  impact of traditional forms of bullying.
\newblock \emph{Journal of youth and adolescence}, 42(5):685--697.

\bibitem[{\c{C}\"{o}ltekin(2020)}]{coltekin2020}
\c{C}a\u{g}r{\i} \c{C}\"{o}ltekin. 2020.
\newblock {A Corpus of Turkish Offensive Language on Social Media}.
\newblock In \emph{Proceedings of LREC}.

\bibitem[{Chiril et~al.(2019)Chiril, Benamara~Zitoune, Moriceau, Coulomb-Gully,
  and Kumar}]{chiril-etal-2019-multilingual}
Patricia Chiril, Farah Benamara~Zitoune, V{\'e}ronique Moriceau, Marl{\`e}ne
  Coulomb-Gully, and Abhishek Kumar. 2019.
\newblock Multilingual and multitarget hate speech detection in tweets.
\newblock In \emph{Proceedings of TALN}.

\bibitem[{Conneau et~al.(2019)Conneau, Khandelwal, Goyal, Chaudhary, Wenzek,
  Guzm{\'a}n, Grave, Ott, Zettlemoyer, and Stoyanov}]{conneau2019unsupervised}
Alexis Conneau, Kartikay Khandelwal, Naman Goyal, Vishrav Chaudhary, Guillaume
  Wenzek, Francisco Guzm{\'a}n, Edouard Grave, Myle Ott, Luke Zettlemoyer, and
  Veselin Stoyanov. 2019.
\newblock Unsupervised cross-lingual representation learning at scale.
\newblock \emph{arXiv preprint arXiv:1911.02116}.

\bibitem[{Dadvar et~al.(2013)Dadvar, Trieschnigg, Ordelman, and
  de~Jong}]{dadvar2013improving}
Maral Dadvar, Dolf Trieschnigg, Roeland Ordelman, and Franciska de~Jong. 2013.
\newblock {Improving Dyberbullying Detection with User Context}.
\newblock In \emph{Proceedings of ECIR}.

\bibitem[{Devlin et~al.(2019)Devlin, Chang, Lee, and
  Toutanova}]{devlin2019bert}
Jacob Devlin, Ming-Wei Chang, Kenton Lee, and Kristina Toutanova. 2019.
\newblock Bert: Pre-training of deep bidirectional transformers for language
  understanding.
\newblock In \emph{Proceedings of NAACL}.

\bibitem[{Fi\v{s}er et~al.(2017)Fi\v{s}er, Erjavec, and
  Ljube\v{s}i\'{c}}]{fiser2017}
Darja Fi\v{s}er, Toma\v{z} Erjavec, and Nikola Ljube\v{s}i\'{c}. 2017.
\newblock {Legal Framework, Dataset and Annotation Schema for Socially
  Unacceptable On-line Discourse Practices in Slovene}.
\newblock In \emph{Proceedings ALW}.

\bibitem[{Fortuna et~al.(2019)Fortuna, da~Silva, Wanner, Nunes
  et~al.}]{fortuna2019hierarchically}
Paula Fortuna, Joao~Rocha da~Silva, Leo Wanner, S{\'e}rgio Nunes, et~al. 2019.
\newblock {A Hierarchically-labeled Portuguese Hate Speech Dataset}.
\newblock In \emph{Proceedings of ALW}.

\bibitem[{Ghadery and Moens(2020)}]{ghadery2020liir}
Erfan Ghadery and Marie-Francine Moens. 2020.
\newblock Liir at semeval-2020 task 12: A cross-lingual augmentation approach
  for multilingual offensive language identification.
\newblock \emph{arXiv preprint arXiv:2005.03695}.

\bibitem[{Karthikeyan et~al.(2020)Karthikeyan, Wang, Mayhew, and
  Roth}]{karthikeyan2020cross}
K~Karthikeyan, Zihan Wang, Stephen Mayhew, and Dan Roth. 2020.
\newblock Cross-lingual ability of multilingual bert: An empirical study.
\newblock In \emph{Proceedings of ICLR}.

\bibitem[{Kumar et~al.(2018)Kumar, Ojha, Malmasi, and
  Zampieri}]{kumar2018benchmarking}
Ritesh Kumar, Atul~Kr Ojha, Shervin Malmasi, and Marcos Zampieri. 2018.
\newblock Benchmarking aggression identification in social media.
\newblock In \emph{Proceedings of TRAC}.

\bibitem[{Kumar et~al.(2020)Kumar, Ojha, Malmasi, and Zampieri}]{trac2020}
Ritesh Kumar, Atul~Kr. Ojha, Shervin Malmasi, and Marcos Zampieri. 2020.
\newblock {Evaluating Aggression Identification in Social Media}.
\newblock In \emph{Proceedings of TRAC}.

\bibitem[{Liu et~al.(2019)Liu, Li, and Zou}]{liu-etal-2019-nuli}
Ping Liu, Wen Li, and Liang Zou. 2019.
\newblock {NULI} at {S}em{E}val-2019 task 6: Transfer learning for offensive
  language detection using bidirectional transformers.
\newblock In \emph{Proceedings of SemEval}.

\bibitem[{Majumder et~al.(2018)Majumder, Mandl et~al.}]{majumder2018filtering}
Prasenjit Majumder, Thomas Mandl, et~al. 2018.
\newblock {Filtering Aggression from the Multilingual Social Media Feed}.
\newblock In \emph{Proceedings TRAC}.

\bibitem[{Malmasi and Zampieri(2017)}]{malmasi2017detecting}
Shervin Malmasi and Marcos Zampieri. 2017.
\newblock {Detecting Hate Speech in Social Media}.
\newblock In \emph{Proceedings of RANLP}.

\bibitem[{Malmasi and Zampieri(2018)}]{malmasi2018}
Shervin Malmasi and Marcos Zampieri. 2018.
\newblock {Challenges in Discriminating Profanity from Hate Speech}.
\newblock \emph{Journal of Experimental \& Theoretical Artificial
  Intelligence}, 30:1 -- 16.

\bibitem[{Mandl et~al.(2019)Mandl, Modha, Majumder, Patel, Dave, Mandlia, and
  Patel}]{hasoc2019}
Thomas Mandl, Sandip Modha, Prasenjit Majumder, Daksh Patel, Mohana Dave,
  Chintak Mandlia, and Aditya Patel. 2019.
\newblock Overview of the hasoc track at fire 2019: Hate speech and offensive
  content identification in indo-european languages.
\newblock In \emph{Proceedings of FIRE}.

\bibitem[{Mubarak et~al.(2017)Mubarak, Kareem, and Walid}]{mubarak2017}
Hamdy Mubarak, Darwish Kareem, and Magdy Walid. 2017.
\newblock Abusive language detection on {A}rabic social media.
\newblock In \emph{Proceedings of ALW}.

\bibitem[{Mubarak et~al.(2020)Mubarak, Rashed, Darwish, Samih, and
  Abdelali}]{mubarak2020arabic}
Hamdy Mubarak, Ammar Rashed, Kareem Darwish, Younes Samih, and Ahmed Abdelali.
  2020.
\newblock Arabic offensive language on twitter: Analysis and experiments.
\newblock \emph{arXiv preprint arXiv:2004.02192}.

\bibitem[{Pamungkas and Patti(2019)}]{pamungkas2019cross}
Endang~Wahyu Pamungkas and Viviana Patti. 2019.
\newblock Cross-domain and cross-lingual abusive language detection: A hybrid
  approach with deep learning and a multilingual lexicon.
\newblock In \emph{Proceedings ACL:SRW}.

\bibitem[{P{\'e}rez and Luque(2019)}]{perez2019atalaya}
Juan~Manuel P{\'e}rez and Franco~M Luque. 2019.
\newblock Atalaya at semeval 2019 task 5: Robust embeddings for tweet
  classification.
\newblock In \emph{Proceedings of SemEval}.

\bibitem[{Peters et~al.(2018)Peters, Neumann, Iyyer, Gardner, Clark, Lee, and
  Zettlemoyer}]{peters-etal-2018-deep}
Matthew Peters, Mark Neumann, Mohit Iyyer, Matt Gardner, Christopher Clark,
  Kenton Lee, and Luke Zettlemoyer. 2018.
\newblock {Deep Contextualized Word Representations}.
\newblock In \emph{Proceedings of NAACL}.

\bibitem[{Pires et~al.(2019)Pires, Schlinger, and
  Garrette}]{pires-etal-2019-multilingual}
Telmo Pires, Eva Schlinger, and Dan Garrette. 2019.
\newblock How multilingual is multilingual {BERT}?
\newblock In \emph{Proceedings of ACL}.

\bibitem[{Pitenis et~al.(2020)Pitenis, Zampieri, and Ranasinghe}]{pitenis2020}
Zeses Pitenis, Marcos Zampieri, and Tharindu Ranasinghe. 2020.
\newblock {Offensive Language Identification in Greek}.
\newblock In \emph{Proceedings of LREC}.

\bibitem[{Poletto et~al.(2017)Poletto, Stranisci, Sanguinetti, Patti, and
  Bosco}]{poletto2017hate}
Fabio Poletto, Marco Stranisci, Manuela Sanguinetti, Viviana Patti, and
  Cristina Bosco. 2017.
\newblock {Hate Speech Annotation: Analysis of an Italian Twitter Corpus}.
\newblock In \emph{Proceedings of CLiC-it}.

\bibitem[{Ranasinghe et~al.(2019)Ranasinghe, Zampieri, and
  Hettiarachchi}]{ranasinghe2019brums}
Tharindu Ranasinghe, Marcos Zampieri, and Hansi Hettiarachchi. 2019.
\newblock Brums at hasoc 2019: Deep learning models for multilingual hate
  speech and offensive language identification.
\newblock In \emph{Proceedings of FIRE}.

\bibitem[{Ridenhour et~al.(2020)Ridenhour, Bagavathi, Raisi, and
  Krishnan}]{ridenhour2020detecting}
Michael Ridenhour, Arunkumar Bagavathi, Elaheh Raisi, and Siddharth Krishnan.
  2020.
\newblock {Detecting Online Hate Speech: Approaches Using Weak Supervision and
  Network Embedding Models}.
\newblock \emph{arXiv preprint arXiv:2007.12724}.

\bibitem[{Risch and Krestel(2020)}]{risch2020bagging}
Julian Risch and Ralf Krestel. 2020.
\newblock Bagging bert models for robust aggression identification.
\newblock In \emph{Proceedings of TRAC}.

\bibitem[{Rosa et~al.(2019)Rosa, Pereira, Ribeiro, Ferreira, Carvalho,
  Oliveira, Coheur, Paulino, Sim{\~a}o, and Trancoso}]{rosa2019automatic}
Hugo Rosa, N~Pereira, Ricardo Ribeiro, Paula~Costa Ferreira, Joao~Paulo
  Carvalho, S~Oliveira, Lu{\'\i}sa Coheur, Paula Paulino, AM~Veiga Sim{\~a}o,
  and Isabel Trancoso. 2019.
\newblock Automatic cyberbullying detection: A systematic review.
\newblock \emph{Computers in Human Behavior}, 93:333--345.

\bibitem[{Rosenthal et~al.(2020)Rosenthal, Atanasova, Karadzhov, Zampieri, and
  Nakov}]{rosenthal2020}
Sara Rosenthal, Pepa Atanasova, Georgi Karadzhov, Marcos Zampieri, and Preslav
  Nakov. 2020.
\newblock {A Large-Scale Weakly Supervised Dataset for Offensive Language
  Identification}.
\newblock In \emph{arXiv preprint arXiv:2004.14454}.

\bibitem[{Sigurbergsson and Derczynski(2020)}]{sigurbergsson2020offensive}
Gudbjartur~Ingi Sigurbergsson and Leon Derczynski. 2020.
\newblock {Offensive Language and Hate Speech Detection for Danish}.
\newblock In \emph{Proceedings of LREC}.

\bibitem[{Sun et~al.(2019)Sun, Qiu, Xu, and
  Huang}]{10.1007/978-3-030-32381-3_16}
Chi Sun, Xipeng Qiu, Yige Xu, and Xuanjing Huang. 2019.
\newblock How to fine-tune bert for text classification?
\newblock In \emph{Chinese Computational Linguistics}, pages 194--206.

\bibitem[{Tulkens et~al.(2016)Tulkens, Hilte, Lodewyckx, Verhoeven, and
  Daelemans}]{tulkens2016dictionary}
St{\'e}phan Tulkens, Lisa Hilte, Elise Lodewyckx, Ben Verhoeven, and Walter
  Daelemans. 2016.
\newblock {A Dictionary-based Approach to Racism Detection in Dutch Social
  Media}.
\newblock In \emph{Proceedings of TA-COS}.

\bibitem[{Vega et~al.(2019)Vega, Reyes-Maga{\~n}a, G{\'o}mez-Adorno, and
  Bel-Enguix}]{vega2019mineriaunam}
Luis Enrique~Argota Vega, Jorge~Carlos Reyes-Maga{\~n}a, Helena
  G{\'o}mez-Adorno, and Gemma Bel-Enguix. 2019.
\newblock Mineriaunam at semeval-2019 task 5: Detecting hate speech in twitter
  using multiple features in a combinatorial framework.
\newblock In \emph{Proceedings of SemEval}.

\bibitem[{Waseem et~al.(2017)Waseem, Davidson, Warmsley, and
  Weber}]{waseem2017understanding}
Zeerak Waseem, Thomas Davidson, Dana Warmsley, and Ingmar Weber. 2017.
\newblock {Understanding Abuse: A Typology of Abusive Language Detection
  Subtasks}.
\newblock In \emph{Proceedings of ALW}.

\bibitem[{Xu et~al.(2012)Xu, Jun, Zhu, and Bellmore}]{xu2012learning}
Jun-Ming Xu, Kwang-Sung Jun, Xiaojin Zhu, and Amy Bellmore. 2012.
\newblock Learning from bullying traces in social media.
\newblock In \emph{Proceedings of NAACL}.

\bibitem[{Yao et~al.(2019)Yao, Chelmis, and Zois}]{yao2019cyberbullying}
Mengfan Yao, Charalampos Chelmis, and Daphney-Stavroula Zois. 2019.
\newblock {Cyberbullying Ends Here: Towards Robust Detection of Cyberbullying
  in Social Media}.
\newblock In \emph{Proceedings of WWW}.

\bibitem[{Zampieri et~al.(2019{\natexlab{a}})Zampieri, Malmasi, Nakov,
  Rosenthal, Farra, and Kumar}]{OLID}
Marcos Zampieri, Shervin Malmasi, Preslav Nakov, Sara Rosenthal, Noura Farra,
  and Ritesh Kumar. 2019{\natexlab{a}}.
\newblock Predicting the type and target of offensive posts in social media.
\newblock In \emph{Proceedings of NAACL}.

\bibitem[{Zampieri et~al.(2019{\natexlab{b}})Zampieri, Malmasi, Nakov,
  Rosenthal, Farra, and Kumar}]{offenseval}
Marcos Zampieri, Shervin Malmasi, Preslav Nakov, Sara Rosenthal, Noura Farra,
  and Ritesh Kumar. 2019{\natexlab{b}}.
\newblock {SemEval-2019 Task 6: Identifying and Categorizing Offensive Language
  in Social Media (OffensEval)}.
\newblock In \emph{Proceedings of SemEval}.

\bibitem[{Zampieri et~al.(2020)Zampieri, Nakov, Rosenthal, Atanasova,
  Karadzhov, Mubarak, Derczynski, Pitenis, and
  C{\"o}ltekin}]{zampieri2020semeval}
Marcos Zampieri, Preslav Nakov, Sara Rosenthal, Pepa Atanasova, Georgi
  Karadzhov, Hamdy Mubarak, Leon Derczynski, Zeses Pitenis, and Cagr{\i}
  C{\"o}ltekin. 2020.
\newblock Semeval-2020 task 12: Multilingual offensive language identification
  in social media (offenseval 2020).
\newblock \emph{Proceedings of SemEval}.

\end{thebibliography}

\end{document}